\documentclass[letterpaper, 10 pt, conference]{ieeeconf}  % Comment this line out if you need a4paper

\IEEEoverridecommandlockouts                              % This command is only needed if 
\usepackage{cite}
\usepackage{enumerate}
\usepackage{graphicx}
\usepackage{epsfig}
\usepackage{color}
\usepackage{threeparttable}
\usepackage{epstopdf}
\usepackage{multirow}
\usepackage{float}
\usepackage{bm}
\usepackage{verbatim}
\usepackage{makecell}
\usepackage{amsmath}

\usepackage{times}
\usepackage{array}
\usepackage{amssymb}
\usepackage{soul}
\usepackage{booktabs}
\usepackage{setspace}
\usepackage{tabularx}
\usepackage{diagbox} 
\usepackage{textcomp}
\usepackage{stfloats}
\usepackage{url}
\usepackage{amsfonts}
\usepackage{hyperref}
\usepackage{xcolor}
\usepackage{colortbl}
\usepackage{pifont}
\usepackage{subfigure}
\usepackage{balance}

\newcommand{\cmark}{\ding{51}}  % 对勾
\newcommand{\xmark}{\ding{55}}  % 叉号

\title{\LARGE \bf
EPRecon: An Efficient Framework for Real-Time Panoptic 3D Reconstruction from Monocular Video
}

\author{Zhen Zhou$^{1, 2}$, Yunkai Ma$^{1, 2, *}$, Junfeng Fan$^{1}$, Shaolin Zhang$^{1}$, Fengshui Jing$^{1, 2}$, and Min Tan$^{1, 2}$% <-this % stops a space
\thanks{This work was supported by the National Natural Science Foundation of China (62173327, 62373354), the Beijing Natural Science Foundation (4232057), and the Youth Innovation Promotion Association of CAS (2022130).}%
\thanks{$^{1}$The authors are with the State Key Laboratory of Multimodal Artificial Intelligence Systems, Institute of Automation, Chinese Academy of Sciences, Beijing, 100190, China. $^{2}$The authors are with the School of Artificial Intelligence, University of Chinese Academy of Sciences, Beijing, 100049, China. * Corresponding author. {\tt\small yunkai.ma@ia.ac.cn}}%
}

\begin{document}

\maketitle
\thispagestyle{empty}
\pagestyle{empty}

%%%%%%%%%%%%%%%%%%%%%%%%%%%%%%%%%%%%%%%%%%%%%%%%%%%%%%%%%%%%%%%%%%%%%%%%%%%%%%%%
\begin{abstract}

Panoptic 3D reconstruction from a monocular video is a fundamental perceptual task in robotic scene understanding. However, existing efforts suffer from inefficiency in terms of inference speed and accuracy, limiting their practical applicability. We present EPRecon, an efficient real-time panoptic 3D reconstruction framework. Current volumetric-based reconstruction methods usually utilize multi-view depth map fusion to obtain scene depth priors, which is time-consuming and poses challenges to real-time scene reconstruction. To address this issue, we propose a lightweight module to directly estimate scene depth priors in a 3D volume for reconstruction quality improvement by generating occupancy probabilities of all voxels. In addition, compared with existing panoptic segmentation methods, EPRecon extracts panoptic features from both voxel features and corresponding image features, obtaining more detailed and comprehensive instance-level semantic information and achieving more accurate segmentation results. Experimental results on the ScanNetV2 dataset demonstrate the superiority of EPRecon over current state-of-the-art methods in terms of both panoptic 3D reconstruction quality and real-time inference. Code is available at \href{https://github.com/zhen6618/EPRecon}{https://github.com/zhen6618/EPRecon}.

\end{abstract}

%%%%%%%%%%%%%%%%%%%%%%%%%%%%%%%%%%%%%%%%%%%%%%%%%%%%%%%%%%%%%%%%%%%%%%%%%%%%%%%%
\section{INTRODUCTION}

Recently, visual-based panoptic 3D reconstruction \cite{PanoRecon} has gradually attracted attention due to its rich scene understanding information and wide application scenarios, such as robotic mapping and manipulation \cite{Robot_Mani}, and augmented reality (AR) \cite{PanopticFusion}. Since camera poses can be accurately obtained by systems such as visual odometry \cite{Visual_Odometry}, robot kinematics \cite{hand_eye_calib}, or Simultaneous Localization and Mapping (SLAM) \cite{ORB_SLAM3, Orbeez-SLAM}, the main challenge in panoptic 3D reconstruction from a monocular video is how to obtain dense, high-quality scene surface depth from continuous RGB sequences and perform panoptic segmentation on the reconstructed scene.

Methods that fuse multi-view depth maps obtained from similarity matching along epipolar lines, such as MVSNet series \cite{MVSNet, MVDepthNet, R-MVSNet, Cascade-MVSNet, Fast-MVSNet}, suffer from depth estimation inconsistencies and redundant computations. To address this issue, volumetric-based methods \cite{Atlas, NeuralRecon, TransformerFusion, VoxFormer, PanoRecon} directly predict the explicit surface occupancy probabilities or implicit truncated signed distance function (TSDF) values to reconstruct scene surfaces in a 3D volume. To improve reconstruction quality, some works \cite{DG_Recon, FineRecon, PanoRecon, Incremental_RAL} first use scene depth estimation models such as multi-view depth map fusion models \cite{SimpleRecon} for a preliminary estimate of scene depth priors, which can eliminate the majority of non-surface voxels (over 90\%), and then perform depth-guided reconstruction. Although these methods effectively improve reconstruction performance, the scene depth estimation networks used in existing methods are time-consuming, posing challenges to real-time scene reconstruction.

To simultaneously obtain panoptic segmentation information and depth information from image sequences, recent studies \cite{PanoRecon, SparseOcc} first predict occupancy probabilities of voxels, and then conduct 3D panoptic segmentation on predicted reconstructed voxels. Although these methods have shown certain efficacy, their panoptic features are only extracted from occupied voxel features or corresponding image features, resulting in limited instance-level semantic information. This is because voxel features need to generate rich geometric features for depth predictions and usually emphasize detailed semantic information. Moreover, after multi-scale feature extraction and aggregation, image features generally emphasize comprehensive semantic information.
  
In this paper, we propose EPRecon, an efficient real-time panoptic 3D reconstruction framework. We first estimate scene depth priors in a 3D volume to eliminate most of non-surface voxels, and then perform depth-guided panoptic 3D reconstruction. Unlike previous works \cite{DG_Recon, FineRecon, PanoRecon, Incremental_RAL} which directly use time-consuming depth estimation models to obtain 2D depth priors and then perform depth fusion, EPRecon introduces a lightweight module to directly estimate scene depth priors in a 3D volume by generating occupancy probabilities of all voxels. Specifically, EPRecon back-projects multi-view image features into the volume and extracts feature similarities between views. These features are further aggregated and converted into occupancy probabilities of voxels. The module makes EPRecon focus more on voxel feature extraction in regions near scene surfaces and improves the entire panoptic reconstruction quality.

Subsequently, to recover dense and coherent panoptic reconstruction results, EPRecon follows the workflow of previous studies \cite{NeuralRecon, PanoRecon} to predict surface occupancy probabilities and TSDF values in a coarse-to-fine manner. Then, 3D panoptic segmentation is performed on the predicted occupied voxels. Different from current methods \cite{PanoRecon, SparseOcc} that extract panoptic features only from occupied voxel features or corresponding image features, we extract panoptic features from both types of features, obtaining more detailed and comprehensive panoptic segmentation information. We use a deformable cross-attention module \cite{DeformDETR} to fuse them. Voxel features make local segmentation regions more detailed and image features make global understanding more comprehensive. The efficient design of EPRecon makes panoptic 3D reconstruction more coherent and reasonable, while also featuring real-time inference. The main contributions are summarized as follows.
\begin{itemize}
\item We propose EPRecon, an efficient real-time panoptic 3D reconstruction framework, which outperforms state-of-the-art (SOTA) methods in terms of both panoptic 3D reconstruction quality and real-time inference.
\item We introduce a lightweight module to directly estimate scene depth priors in a 3D volume, which is significantly faster than multi-view depth map fusion-based methods and improves panoptic reconstruction quality. 
\item Compared with current panoptic segmentation methods, we extract panoptic features from both voxel features and corresponding image features, obtaining more detailed and comprehensive panoptic segmentation information and achieving better segmentation performance.

\end{itemize}

\section{Related Work}
\subsection{3D Geometry Reconstruction}
3D reconstruction from RGB image sequences has a long research history, and many classic works have emerged, such as MVSNet \cite{MVSNet} and ORB-SLAM3 \cite{ORB_SLAM3}. Multi-view stereo reconstruction methods represented by MVSNet series \cite{MVSNet, MVDepthNet, R-MVSNet, Cascade-MVSNet, Fast-MVSNet} estimate depth map for each view and then perform depth map fusion. These methods greatly improve the accuracy of 3D reconstruction, however, multiple depth estimations from different views suffer from depth estimation inconsistencies and redundant computations, usually generating artifacts and repeated surfaces. Volumetric-based approaches \cite{Atlas, NeuralRecon, TransformerFusion, 3DFormer, CVRecon, VoRTX, OpenOccupancy, VIDAR} directly predict surface occupancy probabilities or TSDF values in a 3D volume. An early work is Atlas \cite{Atlas}, which directly regressed a TSDF volume from a set of posed RGB images. NeuralRecon \cite{NeuralRecon} reconstructed local surfaces represented by sparse TSDF volumes for each video fragment sequentially, and used GRUs to fuse TSDF volumes. To achieve more detailed reconstruction, recent studies \cite{DG_Recon, FineRecon, PanoRecon, Incremental_RAL} first estimate scene depth priors and then perform depth-guided scene reconstruction. DG-Recon \cite{DG_Recon} utilized multi-view depth priors to eliminate many irrelevant non-surface voxels and guide surface reconstruction. FineRecon \cite{FineRecon} combined depth priors and global image features for surface occupancy predictions. PanoRecon \cite{PanoRecon} performed panoptic reconstruction with the guidance of MVS depth estimation. Although a preliminary estimate of the scene depth can recover more accurate surfaces, the depth prior prediction modules in existing methods are time-consuming, posing challenges to real-time panoptic reconstruction. To address this issue, we propose a lightweight module to directly estimate scene depth priors in the 3D volume.

\subsection{3D Panoptic Segmentation}
3D panoptic segmentation is mainly performed on voxel features or point features. Previous studies \cite{PANet, DS-Net, LPS_1, LPS_2, LPS_3, LPS_4, LPS_5} mainly integrate the results of semantic segmentation and instance segmentation branches to obtain the final panoptic segmentation results. For example, PANet \cite{PANet} combined semantic predictions generated by a semantic head and detected instances generated by an instance aggregation module to infer panoptic segmentation predictions. Recently, compared to panoptic segmentation methods based on group voting or clustering \cite{vote_cluster_1, vote_cluster_2, PanoRecon}, mask transformer-based methods \cite{MaskPLS, Mask4Former} demonstrate impressive performance by establishing query decoders to directly generate masks for ``thing'' classes and ``stuff'' classes. MaskPLS \cite{MaskPLS} used multiple decoder layers to process learnable queries and point features for generating masks and semantic classes. Mask4Former \cite{Mask4Former} applied the mask transformer framework to 4D panoptic segmentation. Inspired by the competitive performance of the mask transformer framework, we apply this powerful framework for 3D panoptic segmentation without hand-crafted non-learned association strategies. 

\begin{figure*}[t]
    \centering
    \centerline{\includegraphics[scale=0.565]{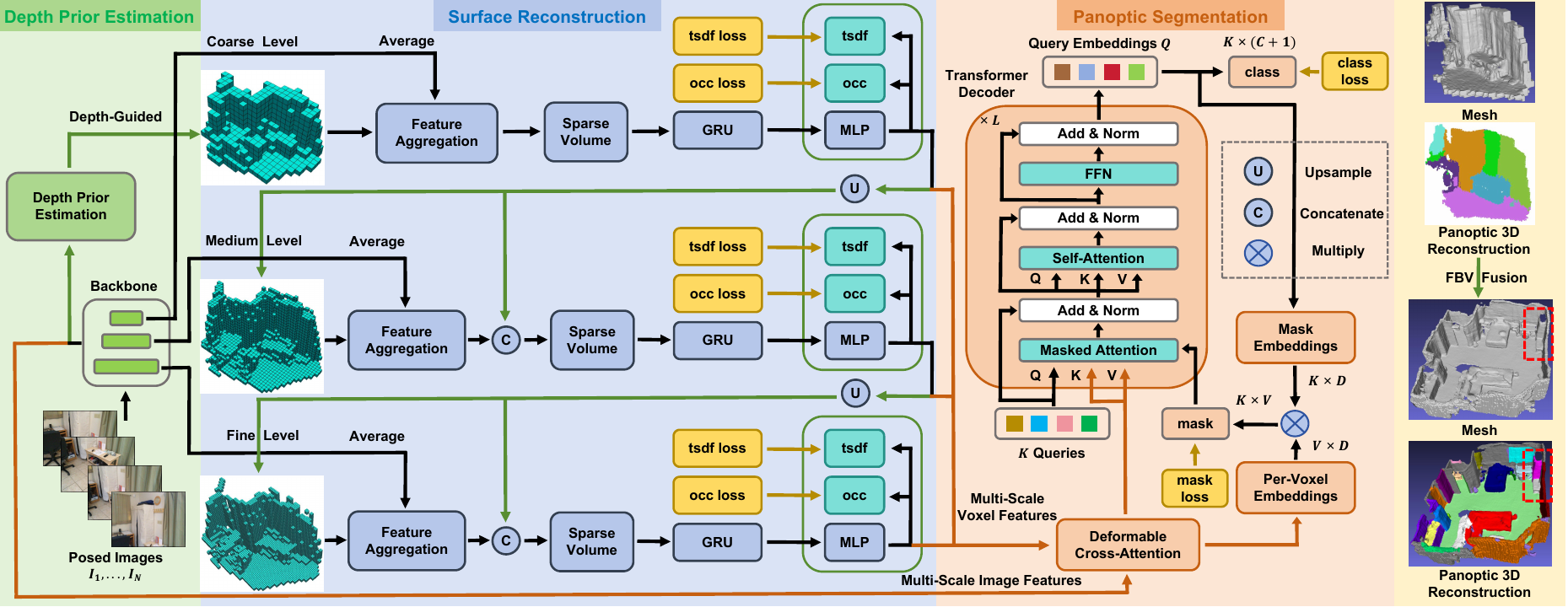}} 
    \caption{Architecture of the proposed EPRecon. EPRecon directly estimates scene depth priors in a 3D volume, and then performs depth-guided panoptic reconstruction. To obtain depth information, we predict the surface occupancy probabilities and TSDF values of voxels. Panoptic features are extracted from both voxel features and corresponding image features, obtaining more detailed and comprehensive instance-level semantic information. EPRecon performs panoptic reconstruction within each FBV and gradually recovers the entire scene in chronological order (see the rightmost column for visualization results).}
    \label{fig:overview}
    \vspace{-0.4cm}
\end{figure*}

\subsection{Panoptic 3D Reconstruction}
A more challenging task is to perform panoptic 3D reconstruction on RGB image sequences. Some video panoptic segmentation approaches such as PVO \cite{PVO}, VPSNet \cite{VPS} and ViP-DeepLab \cite{VIP-DeepLab} can generate depth predictions. With panoptic segmentation, \cite{Fusing_PS_SLAM} predicted panoptic maps based on ORB-SLAM2 \cite{ORB-SLAM2}. However, these methods mainly focus on performing accurate panoptic segmentation on images and exhibit segmentation inconsistencies and limited performance in 3D space. To improve the quality of 3D panoptic segmentation, recent studies \cite{PanoRecon, SparseOcc} directly conduct panoptic segmentation on predicted reconstructed voxels. PanoRecon \cite{PanoRecon} utilized a semantic prediction branch and an offset prediction branch for the predicted occupied voxel features to generate the semantic class and instance center shift vector of each voxel, respectively. Then, the outputs of the two branches were sent to the instance detection module to obtain instance objects. Based on the mask transformer framework \cite{MaskFormer}, SparseOcc \cite{SparseOcc} queried the image features corresponding to predicted occupied voxels and predicted the masks of ``thing'' classes and ``stuff'' classes. Panoptic-FlashOcc \cite{Panoptic-FlashOcc} extracted panoptic features from bird's-eye-view (BEV) voxel features. However, their panoptic features are only extracted from voxel features or corresponding image features, which have limited instance-level semantic information. Our method also uses the mask transformer framework and fuses both types of features, achieving more detailed and comprehensive panoptic segmentation results.

\subsection{Neural Radiance Fields and 3D Gaussian Splatting}
Recently, self-supervised learning-based approaches are developed. Neural Radiance Fields (NeRF) \cite{PanopticNeRF, PNF, PAg-NeRF} series and 3D Gaussian Splatting (3DGS) \cite{CoSSegGaussians} series generate high-quality panoptic segmentation results from novel perspectives while also recovering scene depth information. Although NeRF and 3DGS methods exhibit remarkable performance in scene rendering and understanding, they primarily focus on novel view synthesis, having limited capability in depth information acquisition and usually suffering from noise, artifacts, and missing details. Our proposed method focuses more on recovering high-quality scene depth and is directly supervised by depth information.

\section{Methods}
\subsection{Overview}

Given a sequence of monocular images $\{ I_t \}$  as input, where the intrinsic parameters $\{ K_t \}$ and extrinsic parameters $\{ P_t \}$ are obtained by systems such as SLAM \cite{ORB_SLAM3} systems, EPRecon aims to reconstruct dense and coherent 3D surfaces from these posed images, while simultaneously performing panoptic segmentation on the reconstructed scene surfaces. 

To provide enough motion parallax and keep multi-view co-visibility for panoptic reconstruction, following the training strategy of \cite{NeuralRecon}, a frame is selected as a key frame if its relative translation is greater than $t_{\mathrm{max}}$ and the relative rotation angle is greater than $R_{\mathrm{max}}$. A window with $N$ key frames is defined as a local fragment, and the global panoptic reconstruction result is fused from all local fragments. The maximum visible depth of each view is set to $d_{\mathrm{max}}$, and all view frustums in a fragment are limited to a cubic-shaped and voxelized fragment bounding volume (FBV, $\mathcal{B} \in \mathbb{R}^{B \times B \times B}$). EPRecon is mainly conducted within each FBV, and its overall flowchart is depicted as in Fig. \ref{fig:overview}. It consists of three major stages: depth prior estimation, surface reconstruction, and panoptic segmentation.

\subsection{Depth Prior Estimation}
\begin{figure}[t]
    \centering
    \centerline{\includegraphics[scale=0.400]{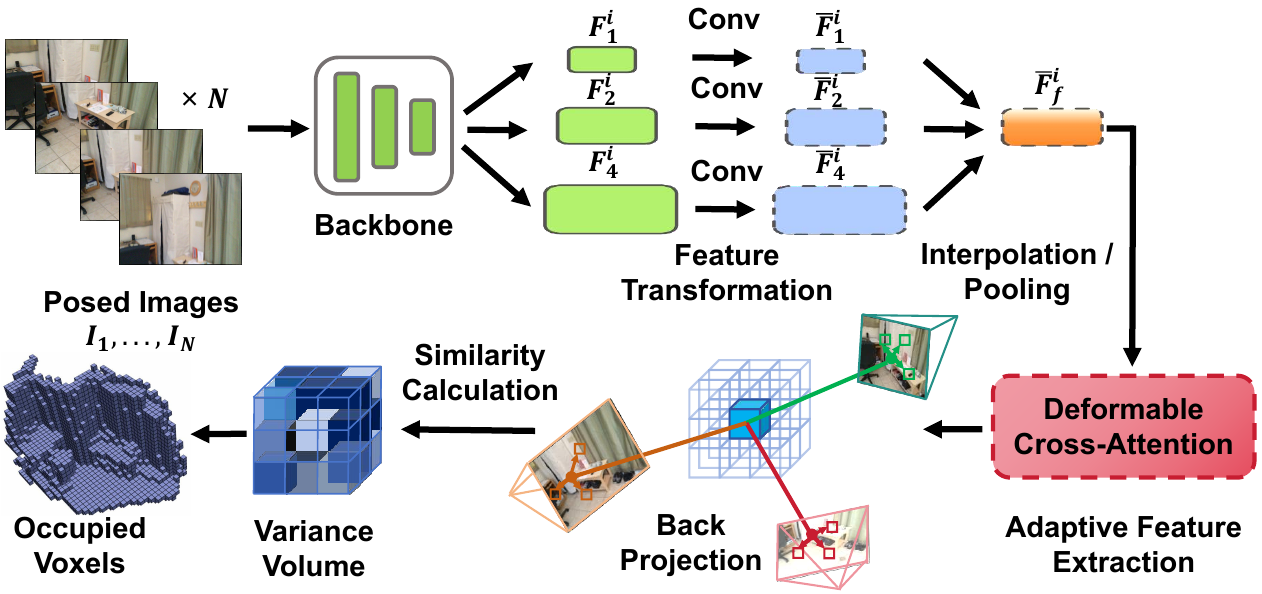}} 
    \caption{Illustration of the proposed depth prior estimation module.}
    \label{fig:init}
    \vspace{-0.6cm}
\end{figure}

EPRecon introduces a lightweight module, which is shown in Fig. \ref{fig:init}, to directly estimate the scene depth priors in the current FBV $\mathcal{B}_t$. Motivated by the principle that the pixel features of each view corresponding to the same voxel in the FBV are similar, we determine whether a voxel is located on scene surfaces based on the similarities between the corresponding pixel features. Given $N$ key frame images $\{ I_1, \ldots, I_N\}$, EPRecon first utilizes a backbone to extract multi-scale image features for each view. The image features at three scales $\{ \mathcal{F}^i_1, \mathcal{F}^i_2, \mathcal{F}^i_4\}$ are extracted for further processing, where the numbers represent relative multiples of feature map resolution and $i$ is the image index. To further highlight the differences between different pixel features, $\{ \mathcal{F}^i_1, \mathcal{F}^i_2, \mathcal{F}^i_4\}$ are converted into features $\{\overline{\mathcal{F}}^i_1, \overline{\mathcal{F}}^i_2, \overline{\mathcal{F}}^i_4\}$ representing pixel similarity via a convolution-based transformation module. Subsequently, we merge $\{\overline{\mathcal{F}}^i_1, \overline{\mathcal{F}}^i_2, \overline{\mathcal{F}}^i_4\}$ into one scale using interpolation and pooling operations, obtaining $\overline{\mathcal{F}}^i_f$. 

After capturing the pixel features of all views, these pixel features are back-projected into the corresponding voxels in the FBV $\mathcal{B}_t$ according to the camera's intrinsic parameters $\{ K_t \}$ and extrinsic parameters $\{ P_t \}$. However, for each voxel, only obtaining the features of one pixel in each view for similarity calculation is not robust enough, and selecting pixel features within a window introduces interference pixels. To address this problem, we adaptively select the features of multiple pixels around each pixel as the features of that pixel, which improves robustness and reduces interference pixels. Specifically, a deformable cross-attention module is applied to adaptively extract pixel features. For each voxel $\mathcal{V}_i$ in the FBV $\mathcal{B}_t$, EPRecon extracts the pixel features $\{p_1, \ldots, p_N\}$ ($p_i \in \mathbb{R}^{C}$, $C$ is the feature dimension) of $N$ corresponding views via back-projection.

Then, we use variance to represent feature similarity across all views. For pixels projected outside the view boundaries, they are not involved in the variance calculation to reduce similarity calculation errors. Hence, the feature mean $\mu_v \in \mathbb{R}^{C}$ of each voxel is calculated as 
\begin{align}
    \mu_v = \frac{m_1 p_1 + m_2 p_2 + \ldots + m_N p_N}{m_1 + m_2 + \ldots + m_N}, \quad m_i \in \{0, 1\},
\end{align}
where $m_i$ indicates whether a voxel $\mathcal{V}_i$ is correctly projected within the view boundary.
The corresponding variance $\sigma^2_v \in \mathbb{R}^{C}$ is formulated as
\begin{align}
    \sigma^2_v = \frac{m_1 (p_1 - \mu_v)^2 + m_2 (p_2 - \mu_v)^2 + \ldots + m_N (p_N - \mu_v)^2}{m_1 + m_2 + \ldots + m_N}.
\end{align}
The variances of all voxels form a variance volume $\Sigma_v$. The advantage of using variance is that its calculation is independent of the number of correctly projected views $\sum m_i$. This allows voxels with fewer correctly projected views to calculate similarities accurately, and voxels with more correctly projected views to enhance the robustness of similarity calculation. 

Subsequently, sparse 3D convolutions are applied to further aggregate the features in the variance volume $\Sigma_v$, followed by a simple Multi-Layer Perceptron (MLP) for transforming variances into surface occupancy probabilities $\mathcal{O}_v$. We use continuous 1-$|$TSDF$|$ ($1-|\mathcal{T}_v|, \mathcal{T}_v \in [-1, 1]$) values as the supervised targets for the predicted surface occupancy probabilities. The log-transformed L1 distance loss is used as the loss function 
\begin{align}
    \mathcal{L}_{D} = \frac{1}{B^3} \sum_{i=1}^{B^3} |\log(\sigma(\mathcal{O}_v) + 1) - \log(2-|\mathcal{T}_v|)|.
\end{align}
Voxels with a surface occupancy probability greater than a predefined threshold are selected as occupied voxels. Then, we select the occupied voxels as the initial voxels for panoptic reconstruction, which has eliminated the majority of non-surface voxels. 
% In addition, to address the hollow and boundary voxels that may occur during the occupancy initialization stage and maintain the completeness between voxels, we then perform 3D morphological optimization on these voxels. Following the morphological processing of 2D pixels, multiple erosions and dilations are applied to the selected occupied voxels.

\subsection{Surface Reconstruction}
Inspired by previous works \cite{NeuralRecon, PanoRecon}, we gradually recover the scene surfaces in a coarse-to-fine manner. Image features from different views are back-projected into occupied voxels, and the average pixel features are taken as the features of the occupied voxels. These voxels with view-agnostic features form a 3D sparse volume, and 3D submanifold sparse convolutions \cite{spconv} are used for further feature extraction, which maintains the geometric structure of the volume.

Subsequently, to maintain temporal coherence and consistency between fragments, the voxel features and corresponding image features of the previous FBV $\mathcal{B}_{t-1}$ are merged into the current FBV $\mathcal{B}_t$. Following \cite{NeuralRecon}, a 3D convolutional variant of the Gated Recurrent Unit (GRU) \cite{GRU} module is applied for temporal fusion.

Finally, surface occupancy probabilities and TSDF values of all voxels in the current volume are predicted via a simple MLP module. Voxels with a surface occupancy probability greater than a predefined threshold are selected as final occupied voxels. Marching Cubes algorithm \cite{Marching_Cubes} is performed on the TSDF values within the predicted occupied voxels to reconstruct the scene mesh and obtain more granular depth.

For supervision, occupancy losses and TSDF losses at all the coarse-to-fine levels have the same definition forms. The occupancy loss $\mathcal{L}_{O}$ is defined as the binary cross-entropy (BCE) loss between the predicted surface occupancy probabilities $\mathcal{O}_r$ and the ground-truth surface occupancy probabilities $\mathcal{O}_r^{gt}$, which is formulated as 
\begin{align}
    \mathcal{L}_{O} = -[\mathcal{O}_r^{gt} \log \sigma(\mathcal{O}_r)
    + (1 - \mathcal{O}_r^{gt}) \log \sigma(1 - \mathcal{O}_r)].
\end{align}
The TSDF loss $\mathcal{L}_{T}$ is defined as the L1 distance loss between the log-transformed predicted TSDF values $\mathcal{T}_r$ and the log-transformed ground-truth TSDF values $\mathcal{T}_r^{gt}$.
\begin{align}
    \mathcal{L}_{T} = |sgn(\mathcal{T}_r) \log(|\mathcal{T}_r| + 1) - sgn(\mathcal{T}_r^{gt}) \log(|\mathcal{T}_r^{gt}| + 1)|,
\end{align}
where $sgn(\cdot)$ is the sign function.

\subsection{Panoptic Segmentation}
Panoptic segmentation is performed on the final predicted occupied voxels. Inspired by the mask transformer framework \cite{MaskFormer, Mask2Former}, EPRecon performs feature queries to occupied voxels, generating a binary mask for each ``thing'' or ``stuff'' class. As voxel features are primarily used for recovering surface geometry, they usually focus more on local geometric structures and more detailed semantic information. Image features excel at capturing more comprehensive semantic information across multiple scales. To recover more detailed and comprehensive panoptic segmentation results, the panoptic features of the occupied voxels are extracted from both voxel features and corresponding image features. To extract richer instance-level semantic information, we first map the occupied voxels predicted at the fine level back to the coarse and medium levels to obtain multi-scale voxel features and corresponding image features. Then, similar to the depth prior estimation module, we use a deformable cross-attention module to fuse voxel features and corresponding image features separately at each scale.

Given $K$ queries and fused multi-scale features $\mathcal{F}_p$, a transformer decoder attends to $\mathcal{F}_p$ and produces $K$ per-segment query embeddings $\mathcal{Q}_p$. In the transformer decoder, there are $L$ repeated transformer blocks, each consisting of a masked cross-attention module, a self-attention module, and a feed-forward network (FFN). Cross-attention associates each query with a ``thing'', ``stuff'' or empty class. Specifically, the masked attention mechanism limits the attention range to the foreground region of the prediction mask of each query, improving interaction efficiency and reducing memory usage. We use Fourier positional encodings \cite{Mask3D} based on voxel positions. Self-attention understands and distinguishes different queries. Subsequently, the $K$ produced query embeddings $\mathcal{Q}_p$ independently generate $K$ class predictions with $K$ corresponding mask embeddings $\mathcal{Q}_p^m \in \mathbb{R}^{K \times D}$ ($D$ is the feature dimension). These mask embeddings $\mathcal{Q}_p^m$ are then dot-producted with per-voxel embeddings $\mathcal{V}_p^m \in \mathbb{R}^{V \times D}$ ($V$ is the number of occupied voxels) extracted from $\mathcal{F}_p$ to generate $K$ binary masks $\mathcal{M}_p^m \in \mathbb{R}^{K \times V}$, where $\mathcal{M}_p^m = \mathcal{Q}_p^m ({\mathcal{V}_p^m})^T$. Then, the predicted masks are used for masked cross-attention in the next transformer block and inference of the final panoptic segmentation results. 

Following \cite{Mask2Former}, an auxiliary loss is added to every intermediate transformer block and to the learnable query features before the transformer decoder. We use BCE loss and dice loss \cite{dice_loss} for the mask loss $\mathcal{L}_{M}$, and cross-entropy (CE) loss for the classification loss $\mathcal{L}_{C}$. The weighted sum of losses from all modules constitutes the final loss
\begin{align}
    \mathcal{L}_{\mathrm{all}} = \lambda_D \mathcal{L}_{D} + \lambda_O \mathcal{L}_{O} + \lambda_T \mathcal{L}_{T} + \lambda_M \mathcal{L}_{M} + \lambda_C \mathcal{L}_{C},
\end{align}
where $\{ \lambda_D, \lambda_O, \lambda_T, \lambda_M, \lambda_C \}$ are weights.

\subsection{Implementation Details}
For the training strategy, we first train the lightweight depth prior estimation module for 24 epochs. Then, we freeze the network weights in the depth prior estimation module, and train the remaining panoptic reconstruction modules for 200 epochs. The entire network is trained on a single NVIDIA Tesla A800 GPU for 14 days. The voxel sizes from coarse to fine levels in the surface reconstruction module are set to 16cm, 8cm, and 4cm, respectively. TSDF truncation distance is 12cm, and $d_{\mathrm{max}}$ is set to 3m. Nearest-neighbor interpolation is used in the upsampling operations between coarse-to-fine levels. $R_{\mathrm{max}}$ and $t_{\mathrm{max}}$ are set to 15° and 0.1m, respectively. We set 9 views as the length of a fragment. We use 6 transformer blocks with 80 queries in the transformer decoder by default. The loss weights $\{ \lambda_D, \lambda_O, \lambda_T, \lambda_M, \lambda_C \}$ are set to $\{ 0.5, 1.0, 1.0, 0.6, 0.24 \}$. For instance fusion across different FBVs, we consider instances with the same semantic label to be the same instance if the IoU (Intersection over Union) between them exceeds 0.05.

\section{Experiments}
\subsection{Dataset}
We conduct panoptic 3D reconstruction experiments on ScanNet(V2) \cite{ScanNet}, which is a large and richly-annotated dataset with RGB images, camera poses, surface reconstruction annotations, and instance-level semantic segmentation annotations. The training set, validation set and test set contain 1201, 312 and 100 scenes, respectively. It contains more than 2,500,000 frames. For fair comparison, we use the official dataset split for training and evaluation. All RGB images are resized to $640 \times 480$.

\subsection{Comparisons with the State-of-the-Art}
We first conduct comparison experiments on the test set to demonstrate the superiority of EPRecon over current SOTA methods. The experimental results of other methods and metric calculation are borrowed from \cite{PanoRecon}.

\subsubsection{3D Geometry Reconstruction} 
We follow the 3D geometry metrics used in \cite{PanoRecon, NeuralRecon} to evaluate the 3D geometry reconstruction performance of our proposed EPRecon. The experimental results are shown in Table \ref{table:3D_geometry}. Our EPRecon is an online geometry reconstruction method and outperforms existing online volumetric-based methods and depth map fusion-based methods in terms of the ``F-score'' metric, which demonstrates that EPRecon has achieved a better balance between the reconstruction accuracy and completeness. Especially, with the assistance of the proposed depth prior estimation module, EPRecon achieves a higher ``precision'' metric than all depth map fusion-based methods and volumetric-based methods, recovering more accurate and concise geometric surfaces. The ``recall'' metric of EPRecon is slightly lower than PanoRecon because EPRecon does not generate excessive redundant geometric structures due to depth map fusion used for depth prior estimation.

\begin{table}[t]
    \caption{Quantitative results of 3D geometry reconstruction. $\dagger$ indicates using depth map fusion. The best results are highlighted in \textcolor{blue}{blue}, \textcolor{orange}{orange}, and \textcolor{purple}{purple}, respectively}
    \footnotesize
    \begin{center}
    \begin{tabular}{m{6.3em}<{\centering}m{2.1em}<{\centering} m{2.1em}<{\centering} m{2.1em}<{\centering} m{2.1em}<{\centering} m{2.1em}<{\centering} m{3.6em}<{\centering}} 

    \hline
    Method & Online & Comp$\downarrow$ & Acc$\downarrow$ & Recall$\uparrow$ & Prec$\uparrow$ & F-score$\uparrow$ \\
    \hline
    MVDNet$\dagger$ \cite{MVDepthNet} & - & \textcolor{blue}{\textbf{4.0}} & 24.0 & \textcolor{blue}{\textbf{0.831}} & 0.208 & 0.329 \\  
    DPSNet$\dagger$ \cite{DPSNet} & - & 4.5 & 28.4 & 0.793 & 0.223 & 0.344 \\     
    GPMVS$\dagger$ \cite{GPMVS} & - & 10.5 & 19.1 & 0.423 & 0.339 & 0.373 \\ 
    SimRec$\dagger$ \cite{SimpleRecon} & - & 6.2 & \textcolor{blue}{\textbf{10.1}} & 0.636 & \textcolor{blue}{\textbf{0.536}} & \textcolor{blue}{\textbf{0.577}} \\

    \hline
    Atlas \cite{Atlas} & \xmark & 8.3 & 10.1 & 0.566 & 0.600 & 0.579 \\
    Vortx \cite{VoRTX} & \xmark & \textcolor{orange}{\textbf{8.1}} & \textcolor{orange}{\textbf{6.2}} & \textcolor{orange}{\textbf{0.605}} & \textcolor{orange}{\textbf{0.689}} & \textcolor{orange}{\textbf{0.643}} \\

    \hline
    NeuRec \cite{NeuralRecon} & \cmark & 13.7 & 5.6 & 0.470 & 0.678 & 0.553 \\
    Zuo et al. \cite{Incremental_RAL} & \cmark & 11.0 & 5.8 & 0.505 & 0.665 & 0.572 \\
    PanoRec \cite{PanoRecon} & \cmark & \textcolor{purple}{\textbf{8.9}} & 6.4 & \textcolor{purple}{\textbf{0.530}} & 0.656 & 0.584 \\

    EPRecon & \cmark & 10.2 & \textcolor{purple}{\textbf{5.1}} & 0.519 & \textcolor{purple}{\textbf{0.692}} & \textcolor{purple}{\textbf{0.593}} \\
    \hline

\end{tabular}
\end{center}
\label{table:3D_geometry}
\end{table}

\begin{table}[t]
    \caption{Quantitative results of 3D semantic segmentation}
    \footnotesize
    \begin{center}
    \begin{tabular}{m{9em}<{\centering} m{6em}<{\centering} m{4em}<{\centering} m{4em}<{\centering}} 

    \hline
    Method & with Depth & Online & mIoU$\uparrow$ \\

    \hline
    ScanNet \cite{ScanNet} & \cmark & - & 30.6 \\
    PointNet++ \cite{PointNet++} & \cmark & - & 33.9 \\
    SPLATNet \cite{SPLATNet} & \cmark & - & 39.3 \\
    3DMV \cite{3DMV} & \cmark & - & 48.4 \\
    SegFusion \cite{SegFusion} & \cmark & - & 51.5 \\
    PanopticFusion \cite{PanopticFusion} & \cmark & - & 52.9 \\
    SparseConvNet \cite{SparseConvNet} & \cmark & - & 72.5 \\
    MinkowskiNet \cite{MinkowskiNet} & \cmark & - & \textbf{73.4} \\

    \hline
    Atlas \cite{Atlas} & \xmark & \xmark & 34.0 \\
    PanoRecon \cite{PanoRecon} & \xmark & \cmark & 52.4 \\
    EPRecon & \xmark & \cmark & \textbf{54.7} \\

    \hline

\end{tabular}
\end{center}
\label{table:3D_semantic}
\vspace{-0.4cm}
\end{table}

\begin{figure*}[t]
    \centering
    \centerline{\includegraphics[scale=0.248]{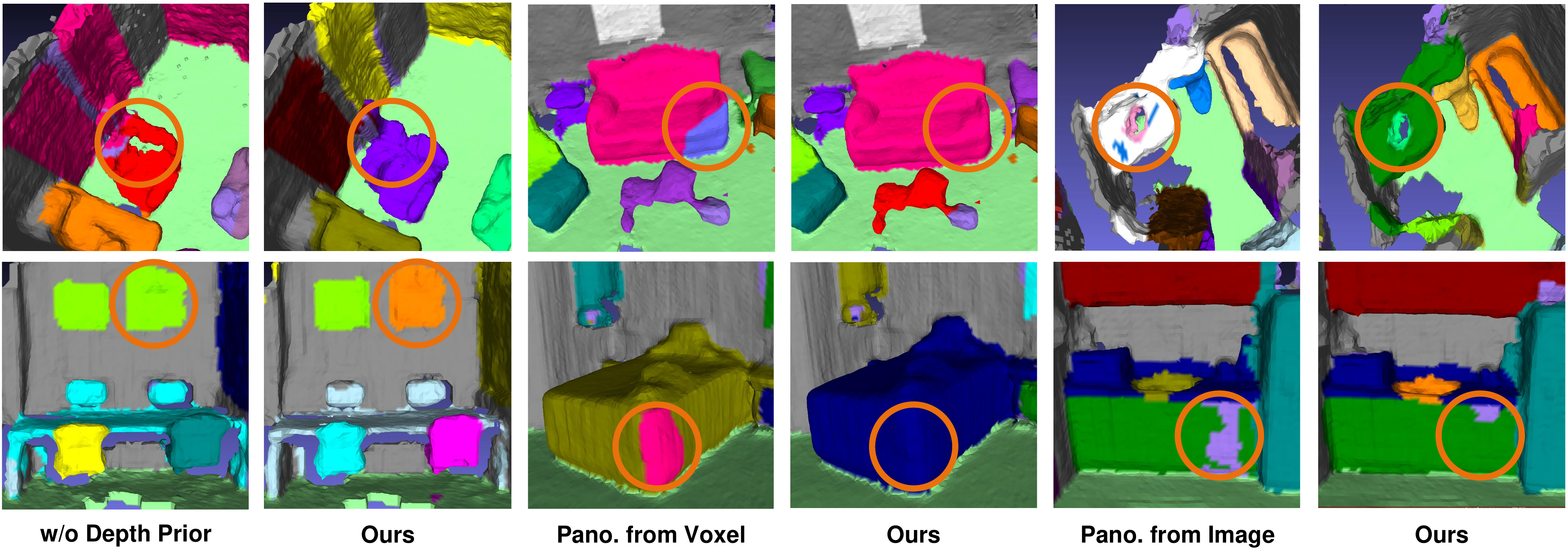}} 
    \caption{Ablation study results of panoptic 3D reconstruction. Under the guidance of the proposed depth prior estimation module, EPRecon recovers more complete and accurate panoptic reconstruction results. Extracting panoptic-related features only from voxel features results in insufficient understanding of more comprehensive instance-level semantic information. Relying only on image features leads to insufficient capture of more detailed information.}
    \label{fig:Ablation_Studies}
    \vspace{-0.2cm}
\end{figure*}

\subsubsection{3D Panoptic Segmentation}
Following \cite{PanoRecon}, we evaluate the 3D semantic segmentation performance in Table \ref{table:3D_semantic} and 3D instance segmentation performance in Table \ref{table:3D_Instance}. For a comprehensive evaluation, we also compare with methods that utilize depth information as input. Despite the unfair setting since we only use RGB data, EPRecon is still surprisingly competitive with (or even outperforms) some previous studies that use 3D input data. Moreover, compared with methods that only take 2D image data as input, our method achieves the best semantic segmentation performance of 54.7\% mIoU, and the best instance segmentation performance of 25.8\% AP50 and 51.6\% AP25. Compared with PanoRecon, which simultaneously generates both geometric and instance-level semantic information from voxel features, EPRecon obtains richer panoptic features from both voxel features and corresponding image features. Furthermore, both PanoRecon and our method are online incremental reconstruction methods and cannot directly obtain global panoptic features, which further demonstrates the superior performance of EPRecon.

\begin{table}[t]
    \caption{Quantitative results of 3D instance segmentation}
    \footnotesize
    \begin{center}
    \begin{tabular}{m{9em}<{\centering} m{5em}<{\centering} m{3em}<{\centering} m{3em}<{\centering} m{3em}<{\centering}} 

    \hline
    Method & with Depth & Online & AP50$\uparrow$ & AP25$\uparrow$ \\

    \hline
    SGPN \cite{SGPN} & \cmark & - & 0.143 & 0.390 \\
    ASIS \cite{ASIS} & \cmark & - & 0.199 & 0.422 \\
    Gspn \cite{Gspn} & \cmark & - & 0.306 & 0.544 \\
    3D-SIS \cite{SGPN} & \cmark & - & 0.382 & 0.558 \\
    SegGroup \cite{SegGroup} & \cmark & - & 0.445 & 0.637 \\
    PBNet \cite{PBNet} & \cmark & - & 0.747 & 0.825 \\
    Mask3D \cite{Mask3D} & \cmark & - & \textbf{0.780} & \textbf{0.870} \\

    \hline
    PanoRecon \cite{PanoRecon} & \xmark & \cmark & 0.227 & 0.484 \\
    EPRecon & \xmark & \cmark & \textbf{0.258} & \textbf{0.516} \\

    \hline

\end{tabular}
\end{center}
\label{table:3D_Instance}
\vspace{-0.4cm}
\end{table}

\subsubsection{Runtime Analysis}
For fair comparison, we perform runtime evaluation on a single NVIDIA RTX3090 GPU. Since PanoRecon first introduces the task of panoptic 3D reconstruction from a posed monocular video, we mainly compare our work with it. As shown in Table \ref{table:Runtime}, each fragment contains 9 key frames, and we give the average inference time for each fragment. Thanks to the lightweight depth prior estimation module, the inference speed of EPRecon is significantly faster than PanoRecon, and EPRecon also achieves higher panoptic 3D reconstruction quality. EPRecon achieves a real-time panoptic 3D reconstruction speed of 31.36 KFPS (key frames per second), which is 2.4$\times$ faster than PanoRecon.

\subsection{Ablation Studies}
In Table \ref{table:Ablation_Studies}, we conduct ablation studies to verify the effectiveness of the components in EPRecon. The qualitative comparison and analysis are shown in Fig. \ref{fig:Ablation_Studies}. The proposed depth prior estimation module effectively improves the quality of 3D geometric reconstruction and 3D panoptic segmentation. Since we directly estimate scene depth priors in a 3D volume, our proposed module is more lightweight (10.8$\times$ faster) than current depth map fusion-based methods used for depth prior estimation (see Table \ref{table:Runtime}). In addition, we compare with extracting panoptic features only from voxel features or corresponding image features. Our method captures both types of features, having more detailed and comprehensive instance-level semantic information, which leads to higher 3D panoptic segmentation performance. 

\begin{table}[t]
    \caption{Runtime analysis of panoptic 3D reconstruction. PanoRecon uses multi-view depth map fusion for depth prior estimation}
    \footnotesize
    \begin{center}
    \begin{tabular}{m{6.3em}<{\centering} m{6.2em}<{\centering} m{6.5em}<{\centering} m{5.5em}<{\centering}} 

    \hline
    Method & Depth Prior Estimation & Panoptic 3D Reconstruction & Runtime \\

    \hline
    PanoRecon \cite{PanoRecon} & 430 ms & 700 ms & 12.85 KFPS \\
    EPRecon & \textbf{40 ms} (10.8$\times$) & \textbf{287 ms} (2.4$\times$) & \textbf{31.36 KFPS} \\

    \hline

\end{tabular}
\end{center}
\label{table:Runtime}
\end{table}

\begin{table}[t]
    \caption{Ablation studies on the ScanNet(V2) validation set}
    \footnotesize
    \begin{center}
    \begin{tabular}{m{7.3em}<{\centering}m{2.0em}<{\centering} m{2.0em}<{\centering} m{3.0em}<{\centering} m{1.9em}<{\centering} m{1.9em}<{\centering} m{1.9em}<{\centering}} 

    \hline
    Method & Recall & Prec & F-score & mIoU & AP50 & AP25 \\
    \hline
    
    w/o Depth Prior & 0.524 & 0.681 & 0.592 & 55.0 & 0.265 & 0.515 \\

    Pano. from Voxel & 0.548 & 0.694 & 0.612 & 55.5 & 0.269 & 0.520 \\

    Pano. from Image & 0.567 & 0.700 & 0.626 & 55.8 & 0.273 & 0.524 \\

    Ours & \textbf{0.575} & \textbf{0.709} & \textbf{0.635} & \textbf{56.3} & \textbf{0.289} & \textbf{0.531} \\
    \hline

\end{tabular}
\end{center}
\label{table:Ablation_Studies}
\vspace{-0.4cm}
\end{table}

\section{Conclusion}
In this work, we propose EPRecon, an efficient real-time panoptic 3D reconstruction framework. In EPRecon, a lightweight module is introduced to directly estimate scene depth priors in a 3D volume, eliminating the majority of non-surface voxels and improving panoptic reconstruction quality. Additionally, we extract panoptic features from both voxel features and corresponding image features, obtaining more detailed and comprehensive panoptic segmentation information. Experimental results demonstrate the superiority of EPRecon over SOTA methods in terms of panoptic reconstruction quality and real-time inference. The limitations are that the model training time is long and the non-learning-based instance fusion method between FBVs is not efficient. Future work will focus on improvements in these aspects.

%                                   % the textheight of the last page by a suitable amount.
%                                   % This command does not take effect until the next page
%                                   % so it should come on the page before the last. Make
%                                   % sure that you do not shorten the textheight too much.
    
\bibliographystyle{IEEEtran}
\bibliography{IEEEabrv, root}

\end{document}